\documentclass[conference]{IEEEtran}
\IEEEoverridecommandlockouts
\usepackage{cite}
\usepackage{booktabs}
\usepackage{amsmath,amssymb,amsfonts}
\usepackage{algorithmic}
\usepackage{graphicx}
\usepackage{textcomp}
\usepackage[dvipsnames]{xcolor}
\usepackage{url}

\def\BibTeX{{\rm B\kern-.05em{\sc i\kern-.025em b}\kern-.08em
    T\kern-.1667em\lower.7ex\hbox{E}\kern-.125emX}}
\begin{document}

\title{Single-Beat Cuffless Blood Pressure Estimation Using Ear-PPG and ECG with a Lightweight Hybrid Learning Framework\\
\thanks{\IEEEauthorrefmark{2}Contribute equally to this work. $^*$Corresponding author. E-mail address: yayun.du@vanderbilt.edu. \\
This work was supported by the Department of Electrical and Computer Engineering and Vanderbilt Institute for Surgery and Engineering, Vanderbilt University, Nashville, TN, USA. }
}

\author{
\IEEEauthorblockN{
Kindeep~K.~Dhatt\IEEEauthorrefmark{2}$^{1,6}$,
Tengyue~Wu\IEEEauthorrefmark{2}$^2$,
Hanbang~Hua$^3$,
Yayun~Du$^{1,2,4,5,6}$\IEEEauthorrefmark{1}
}
\IEEEauthorblockA{1. Vanderbilt Institute for Surgery and Engineering, Vanderbilt University, Nashville, TN, 37232}
\IEEEauthorblockA{2. Department of Electrical and Computer Engineering, Vanderbilt University, Nashville, TN, 37235}
\IEEEauthorblockA{3. Querrey Simpson Institute for Bioelectronics, Northwestern University, Evanston, IL 60208}
\IEEEauthorblockA{4. Department of Computer Science, Vanderbilt University, Nashville, TN, 37212}
\IEEEauthorblockA{5. Department of Mechanical Engineering, Vanderbilt University, Nashville, TN, 37240}
\IEEEauthorblockA{6. Department of Biomedical Engineering, Vanderbilt University, Nashville, TN, 37235}
}

\maketitle

\begin{abstract}
Continuous cuffless blood pressure (BP) monitoring remains challenging due to motion artifacts, physiological variability, and the limited robustness of conventional pulse transit time (PTT) models under dynamic conditions. Many prior approaches rely on multi-second windows to stabilize estimation, an assumption that is frequently violated during real-world monitoring with intermittent signal corruption. Here, we show that discriminative BP-related information is preserved at the single-beat level and present a lightweight multi-modal wearable framework for continuous BP estimation. The system integrates synchronized chest electrocardiography (ECG) and ear-clip reflectance photoplethysmography, each co-located with a 6-axis inertial measurement unit to provide motion context. We introduce a hybrid learning architecture in which a one-dimensional convolutional neural network extracts a 64-dimensional embedding from individual PPG beats and fuses it with 30 physiology-grounded features, including PTT statistics and heart rate variability, followed by LightGBM regression. The method was evaluated using a multi-phase stress protocol ($n=10$) and the PulseDB public dataset with subject-disjoint validation. Across 30 independent runs, the model achieved mean absolute errors of $4.02 \pm 0.21$~mmHg for systolic BP and $1.79 \pm 0.05$~mmHg for diastolic BP, corresponding to a 28.2\% reduction in combined MAE relative to baseline models. By enabling beat-wise estimation without long temporal context, this framework supports computationally efficient cuffless BP monitoring suitable for wearable deployment under practical resource constraints. The source code for this work is available at \url{https://github.com/SYMBIOX-Lab/BP-wireless}.

\end{abstract}


\section{Introduction}
Hypertension is a major global health challenge, affecting approximately one in three adults and contributing to $\sim$12.5\% of worldwide deaths \cite{10.36835/dnursing.v1i1.110}. The World Health Organization estimates that over 1.28 billion adults aged 30--79 live with hypertension, but fewer than half achieve adequate blood pressure (BP) control \cite{who_hyp}. In the United States, only $\sim$50\% of diagnosed individuals effectively manage their condition \cite{10.1155/2022/1962475}. Because hypertension is often asymptomatic and episodic, inadequate monitoring can delay diagnosis and intervention, increasing the risk of myocardial infarction, stroke, and other cardiovascular events. Effective BP management therefore requires monitoring across daily activities and transient physiological stressors, rather than isolated resting measurements.

Current clinical practice relies primarily on cuff-based BP measurements, which remain the gold standard but are poorly suited for frequent or continuous use. Cuff devices provide discontinuous measurements, require active user engagement, and exhibit limited adherence in longitudinal monitoring. Nearly 50\% of patients fail to reach BP targets due to poor adherence \cite{burnier_adherence}, and consumer-grade home cuffs have been reported to exhibit biases exceeding 10~mmHg and reduced sensitivity for systolic hypertension \cite{zhou202410}. Repeated cuff inflation may also cause discomfort, skin irritation, or nerve compression, further limiting compliance.

These limitations have motivated growing interest in cuffless BP estimation using photoplethysmography (PPG) and pulse transit time (PTT). However, translation to robust real-world monitoring remains challenging due to three key limitations.
\emph{First}, PPG-based systems are highly sensitive to motion and posture-induced hemodynamic variability. Finger- and wrist-worn sensors are particularly vulnerable to motion artifacts, hydrostatic pressure shifts, and activity-dependent changes in vascular tone, which confound true BP-related dynamics. Linear PTT models often degrade when BP deviates by more than $\sim$10~mmHg or when autonomic state changes due to thermal or metabolic stress \cite{10.1109/tbme.2017.2756018}. Many commercial devices lack integrated inertial sensing, limiting their ability to separate physiological variation from motion-induced confounds; for example, hydrostatic pressure errors of $\sim$7~mmHg per 10~cm of vertical displacement can distort wrist-based measurements \cite{10.1109/tbme.2016.2580904}. \emph{Second}, many BP estimation algorithms rely on long temporal contexts to stabilize feature extraction, assuming multi-second windows or multiple consecutive high-quality PPG beats \cite{haddad2021continuous, long2023bpnet, suhas2024endtoend}. During continuous monitoring, motion, perspiration, and intermittent loss of skin contact can corrupt individual beats, such that a single degraded beat may invalidate an entire estimation window. \emph{Third}, recent advances increasingly depend on computationally intensive deep learning models. Long-sequence processing and deep temporal architectures impose power and latency costs that conflict with the energy and memory constraints of wearable devices \cite{10.3390/s21051595}.

To address these challenges, we propose a lightweight multi-modal framework for continuous BP estimation under dynamic physiological conditions. We introduce a wearable system integrating synchronized chest electrocardiography (ECG) and ear-clip reflectance PPG sensors, each instrumented with a 6-axis inertial measurement unit (IMU). By placing the optical sensor on the earlobe---an anatomically anchored site less susceptible to limb-induced motion than the wrist or finger---and leveraging IMU context, the proposed hardware reduces motion- and posture-related confounds at acquisition. At the algorithmic level, we combine (i) a one-dimensional convolutional neural network (CNN) that extracts beat-level morphological embeddings from individual PPG beats and (ii) physiology-grounded features, including PTT and heart rate variability (HRV), fused within a LightGBM regressor. This design avoids long temporal windows and heavy sequence models while remaining suitable for low-power deployment. The primary contributions are:

\begin{itemize}
    \item \textbf{Robust Multi-Modal Sensing Platform:} We present a synchronized ECG and ear-PPG wearable system with dual IMUs designed to mitigate motion- and posture-induced confounds. The system is evaluated using a multi-phase physiological stress protocol ($n=10$) incorporating rest, exercise, recovery, and cold pressor stress.

    \item \textbf{Single-Beat Blood Pressure Estimation:} We show that BP-relevant morphological information is preserved at the single-beat level, enabling resilient estimation from individual beats without requiring extended temporal context.

    \item \textbf{Lightweight Hybrid Learning Architecture:} We introduce a computationally efficient two-branch model that fuses CNN-derived embeddings with physiological features using a LightGBM regressor \cite{lightgbm}, achieving a 28.2\% reduction in combined MAE relative to baseline models while maintaining a small memory footprint.
\end{itemize}

\section{Methods}
\subsection{Robust Multi-Modal Sensing Platform}
\label{subsec:sensors}
\begin{figure}[htbp]
     \centering
     \includegraphics[width=1\linewidth]{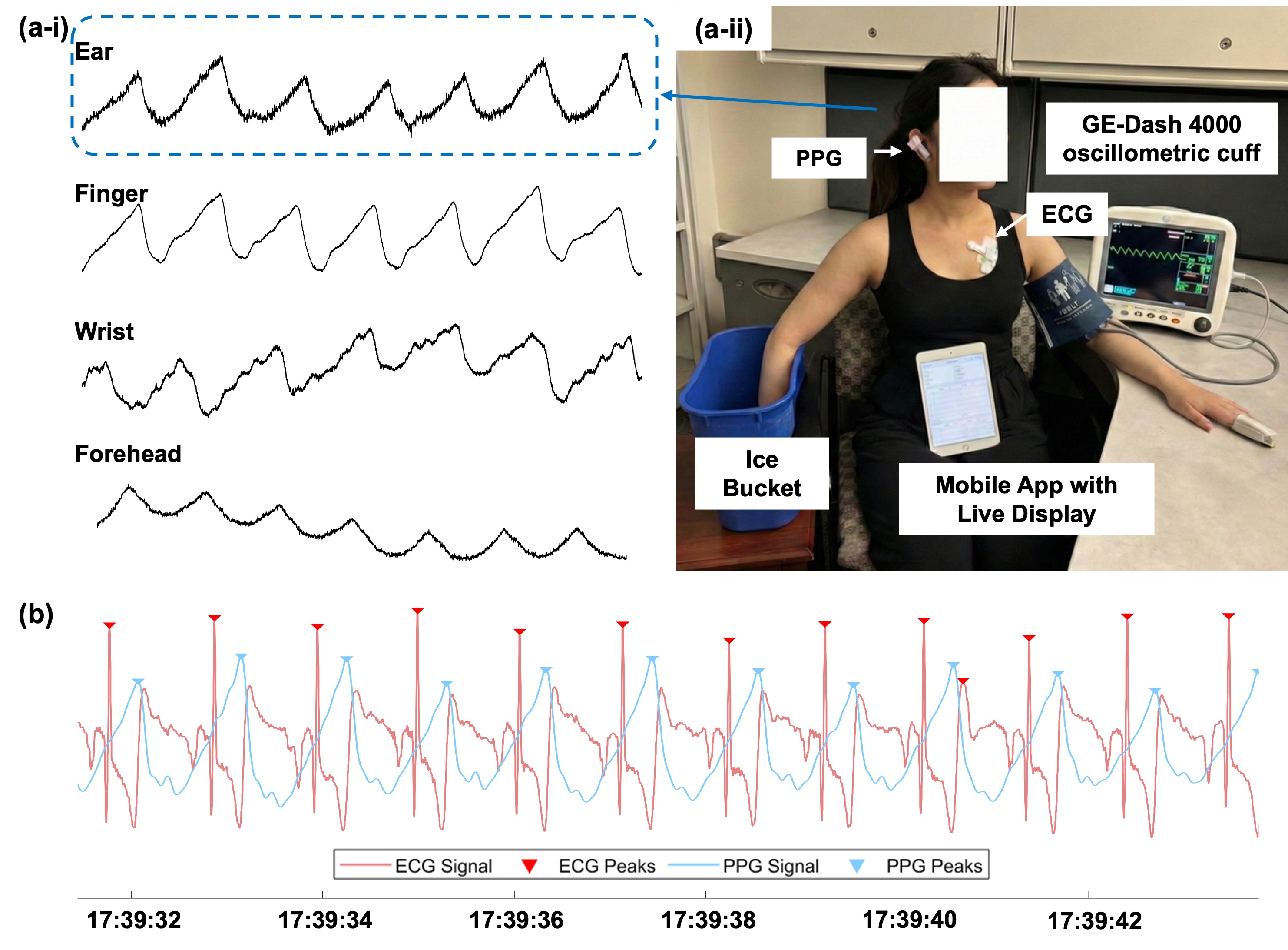}
     \caption{(a) Ear-clip PPG selection and experimental setup. (a-i) Representative PPG waveforms from ear, finger, wrist, and forehead. (a-ii) Wearable acquisition setup with synchronized ECG and ear PPG. (b) Example $\sim$10 s synchronized ECG and ear-PPG with detected peaks; PTT is computed as the delay between paired ECG R-peaks to PPG systolic peaks.The ice-bucket phase served as a cold-pressor stimulus to induce autonomic-mediated peripheral vasoconstriction and transient blood pressure elevation, thereby expanding the physiological BP range for evaluating single-beat cuffless BP measurement.} \label{fig:signal-exp-setup}
 \end{figure}

We developed a synchronized multi-modal wearable sensing platform that integrates chest ECG, ear-clip PPG, and inertial sensing to enable robust cuffless BP estimation under dynamic physiological conditions. The system comprises two time-synchronized wearable sensors (Fig.~\ref{fig:signal-exp-setup}b, Fig.~\ref{fig:sensor-system-arch}) designed to acquire high-fidelity hemodynamic signals during movement. The chest-mounted unit provides a single-lead ECG, positioned near the V2--V4 region with a right-leg drive reference electrode to ensure stable cardiac electrical recordings. ECG signals were sampled at 512~Hz. This module also integrates a 6-axis inertial measurement unit (IMU), sampled at 104~Hz in this study, to capture body dynamics (e.g., posture changes) that introduce hydrostatic pressure offsets.
The PPG module consists of a compact, ultra-flexible near-infrared spectroscopy device clipped to the right earlobe. The sensor employs a dual-wavelength optical configuration (red: 740~nm; infrared: 850~nm) and two photodiodes placed at 12~mm and 17~mm from the LED pair. This configuration yields four raw PPG channels, providing wavelength diversity and depth sensitivity to local tissue perfusion. The optical front-end is driven by a low-noise analog front-end with an 8-bit programmable LED current DAC and a 19-bit sigma--delta ADC, sampling PPG signals at 128~Hz. An identical 6-axis IMU is co-located on the ear module to capture head motion and identify periods of excessive motion that may compromise PPG morphology. Both wearable sensors stream data via Bluetooth Low Energy (BLE) to a local mobile gateway (e.g., an iOS tablet), with a measured inter-device synchronization latency of less than 10~ms. This architecture enables precise temporal alignment of ECG, PPG, and inertial signals for beat-level analysis.


The earlobe was selected as the PPG sensing site due to its superior anatomical stability and robustness under motion. While fingertip PPG provides high signal amplitude, it obstructs manual tasks and is highly susceptible to motion artifacts. Wrist-based measurements suffer from tendon motion, muscle activity, and hydrostatic pressure effects, resulting in pronounced baseline wander. Forehead PPG, although proximal, typically exhibits reduced pulsatile morphology and greater amplitude variability. As shown in Fig.~\ref{fig:signal-exp-setup}a, ear-based PPG maintains a stable baseline and consistent pulse morphology relative to finger-, wrist-, and forehead-based recordings under elevated activity. Additionally, the earlobe is supplied by branches of the carotid artery, preserving perfusion during peripheral vasoconstriction (e.g., cold pressor stress), where distal sites often degrade.


Figure~\ref{fig:signal-exp-setup}b illustrates the temporal alignment between chest ECG and ear-based PPG signals. The physiological delay between ventricular depolarization (ECG R-peak) and the arrival of the arterial pulse at the earlobe (PPG systolic peak) defines the pulse transit time (PTT). PTT was computed on a per-beat basis as $\text{PTT} = t_{\text{PPG\_peak}} - t_{\text{ECG\_Rpeak}}$ and serves as a primary hemodynamic feature for subsequent BP estimation.

\subsection{Data Collection Protocol}

\color{black}
 \begin{figure}[htbp]
     \centering
     \includegraphics[width=1.00\linewidth]{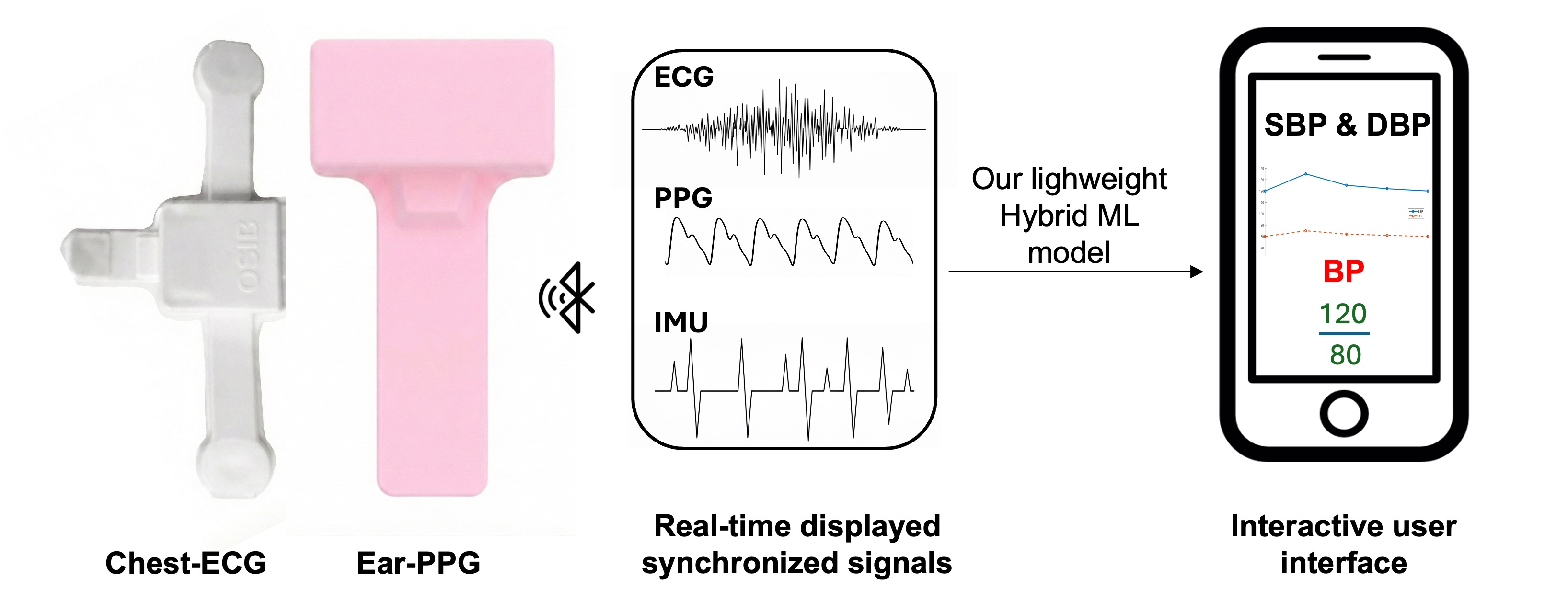}
     \caption{System overview and data flow. Chest ECG and ear-clip PPG (with IMU) are streamed via BLE to a mobile gateway for synchronized visualization and data logging. The proposed lightweight hybrid model estimates SBP and DBP from the recorded signals; this study reports BP from offline processing, while the pipeline is compatible with mobile (e.g., iOS) deployment.}
     \label{fig:sensor-system-arch}
 \end{figure}
During each session, synchronized ECG/PPG/IMU packets were timestamped and buffered on a mobile gateway (iOS tablet) to preserve alignment across modalities (Fig.~\ref{fig:sensor-system-arch}). Recordings were then exported for offline preprocessing and analysis. Given the small model footprint, the same pipeline is compatible with deployment on mobile gateways (e.g., iOS) for near-real-time inference and user feedback; in this study, BP estimates are reported from offline processing.

To ensure stable sensor-skin contact during dynamic movement, commercially available biocompatible medical adhesives (Solventum~2487 and Tegaderm\texttrademark, 3M) were used to reinforce both the chest ECG electrodes and ear-clip PPG sensor. Adhesives were applied without restricting blood flow or causing subject discomfort.

 %
%
\subsection{Study Population and Experimental Protocol}
\label{subsec:protocol}
A pilot feasibility study was conducted on $N=10$ healthy volunteers (5 male, 5 female; Fitzpatrick skin types: I--VI, age: 18-30). The study protocol was approved by the institutional review board (IRB \#212305), and all participants provided written informed consent. To evaluate robustness across diverse autonomic states, participants completed a 20-minute multi-phase protocol adapted from prior cuffless blood pressure validation studies \cite{MILATZ2014e53}, \cite{XiangTing2024ANMP}. The protocol was designed to induce a wide range of blood pressure variations and to decouple PTT from heart rate through distinct physiological mechanisms. The four continuous 5-minute phases were:
\begin{enumerate}
    \item \textbf{Baseline Rest (5 minutes):} Seated rest to establish steady-state hemodynamics.
    \item \textbf{Metabolic Stress (5 minutes):} Moderate-intensity stationary cycling (target heart rate $>100$ bpm) to elevate BP primarily via increased cardiac output. 
    \item \textbf{Recovery (5 minutes):} Post-exercise cool-down period.
    \item \textbf{Thermal Stress (5 minutes):} Cold pressor test with right-hand immersion in ice water ($\sim$4$^{\circ}$C) to induce peripheral vasoconstriction and increased total peripheral resistance, a condition under which linear PTT models are known to degrade. 
\end{enumerate}
Expect the cycling stage, reference blood pressure was acquired using an FDA-cleared clinical oscillometric BP measurements (GE Dash 4000) were obtained at 60-second intervals and temporally aligned with the wearable data stream for validation. Cuff-based measurements were omitted during exercise due to motion artifacts, practical difficulty of collection, and reduced accuracy under dynamic conditions.
\subsection{Datasets and Data Preprocessing}
\textbf{Public Datasets:} We utilized the open-source PulseDB dataset\cite{pulsedb} as a large-scale benchmark for model development. PulseDB provides synchronized ECG, PPG, and arterial blood pressure (ABP) recordings organized at the subject level, enabling subject-disjoint partitioning to prevent information leakage between training and evaluation sets. In this study, data from 100 subjects with at least 5 minutes of continuous ECG, PPG, and ABP recordings were randomly selected to establish a robust baseline model. The dataset represents clinically monitored hospital-based patients rather than healthy individuals. Detailed disease annotations, including arrhythmia status, were not used for subject stratification in this study. Therefore, differences in demographic and clinical characteristics between the PulseDB cohort and our healthy volunteer cohort may affect model generalizability.

\textbf{Our Wearable Dataset:}
Complementary to the public dataset, we collected a wearable dataset using the custom multi-modal sensing platform described in Section~\ref{subsec:sensors}. This dataset captures synchronized ECG, multi-wavelength ear-PPG, and inertial signals under dynamic physiological conditions, including elevated physical activity and thermal stress, which are not represented in ICU-based datasets. Due to its limited sample size ($N=10$), the wearable dataset was used primarily to evaluate model robustness and generalization under motion and autonomic stress rather than for large-scale training.

\textbf{Preprocessing:} ECG, PPG, and reference ABP signals from the public dataset were resampled to a unified sampling frequency of 125~Hz to ensure temporal alignment across modalities. For CNN-based embedding extraction, each segmented PPG beat was resampled to 128 samples to obtain a compact single-beat representation suitable for lightweight inference. Each beat was amplitude-normalized using min--max scaling to reduce inter-subject variability and improve training stability. Beat boundaries were defined using pulse onsets (systolic foot), detected as local minima within a 0.3~s search window preceding each PPG systolic peak. Consecutive onsets were used to segment the continuous PPG waveform into individual cardiac cycles. To enforce physiological plausibility, beats with computed PTT values outside 0.05--0.5~s were excluded from further analysis.

\subsection{Signal Processing and Model Development}
Following beat-level preprocessing, two complementary feature representations were extracted from the synchronized ECG and PPG signals: (i) physiology-grounded hand-crafted features derived from raw ECG and PPG, and (ii) data-driven morphological embeddings learned from single-beat PPG waveforms. This two-branch design enables robust BP estimation while maintaining low input dimensionality and real-time feasibility.

\textbf{Physiology-grounded features:}  
To encode established cardiovascular mechanisms, a set of hand-crafted features was extracted from the ECG and PPG signals. In addition to PTT statistics (mean, median, and standard deviation), we computed heart rate variability (HRV) metrics from sequences of R-R intervals, motivated by prior studies showing significant associations between reduced HRV and elevated BP \cite{schroeder2003hypertension, yugar2023role}. Extracted HRV features included time-domain measures (e.g., mean heart rate, SDNN, RMSSD, pNN50), frequency-domain features (e.g., LF, HF, and LF/HF ratio), and geometric descriptors (e.g., Poincaré plot metrics), and basic statistical summaries. These features capture autonomic nervous system regulation that is not directly observable from a single PPG waveform. Together with selected PPG morphological descriptors, a total of 30 hand-crafted features formed the first input branch of the model.

\begin{figure*}[htbp]
    \centering
    \includegraphics[width=0.8\textwidth]{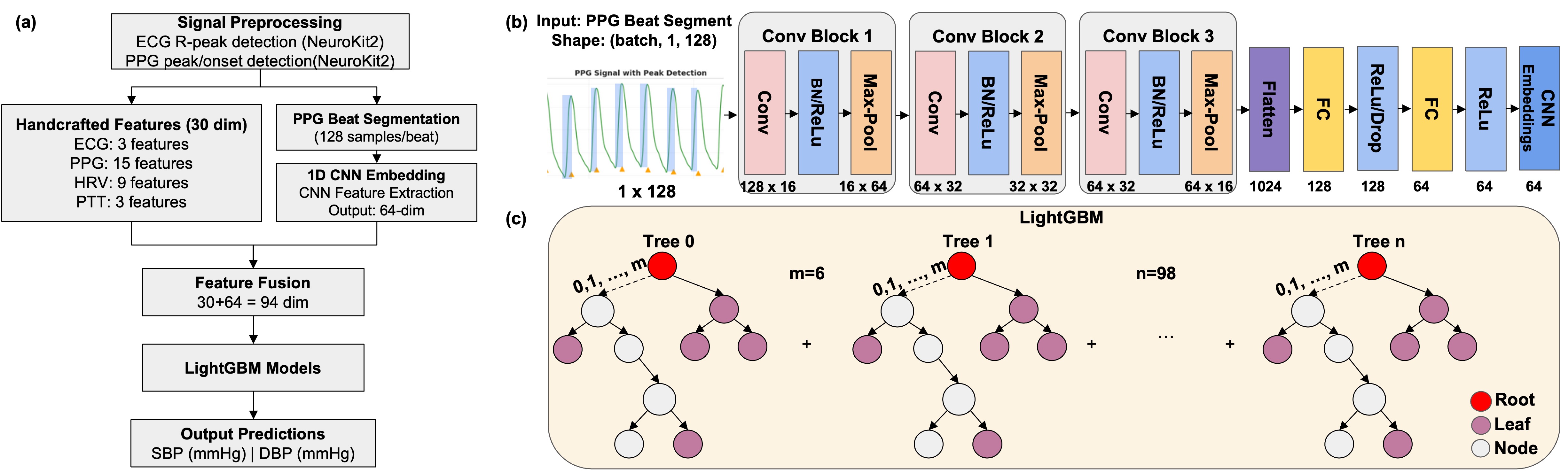}
    \caption{Proposed hybrid CNN+LightGBM framework for cuffless blood pressure estimation. (a) Two-branch feature construction: physiology-grounded features extracted from synchronized ECG/PPG (left) and a CNN-derived embedding from a single normalized PPG beat (right), followed by feature fusion. The fused feature vector is provided to lightGBM regressors to estimate SBP and DBP. (b) One-dimensional CNN feature extractor that maps a 128-sample PPG beat segment to a 64-dimensional morphological embedding. (c) LightGBM regressor that operates on the fused feature vector to estimate SBP and DBP via an ensemble of decision trees. }
    \label{Fig: Model Archtechture}
\end{figure*}

\textbf{PPG morphological embeddings:}  
To capture fine-grained, non-linear waveform characteristics at the single-beat level, a one-dimensional CNN was used as a trainable feature extractor operating solely on PPG signals. As shown in Fig.~\ref{Fig: Model Archtechture}(b), the CNN processes normalized single-beat PPG segments of length 128 samples and consists of three convolutional blocks with channel dimensions $1\!\rightarrow\!16\!\rightarrow\!32\!\rightarrow\!64$ and kernel sizes of 7, 5, and 3. Each block includes batch normalization, ReLU activation, and max pooling to progressively reduce temporal resolution while preserving salient morphology. The final feature maps are flattened and passed through fully connected layers with dropout, producing a compact 64-dimensional embedding for each PPG beat. This branch enables BP estimation from individual cardiac cycles without reliance on long temporal windows.

\textbf{Feature fusion and regression:}  
The final feature is constructed by concatenating the 30-dimensional hand-crafted physiological features with the 64-dimensional CNN-derived PPG embedding, forming a 94-dimensional representation (Fig.~\ref{Fig: Model Archtechture}(c)). This fused feature vector serves as input to a LightGBM regression model to estimate systolic and diastolic blood pressure (SBP and DBP).

LightGBM was selected for its ability to model non-linear feature interactions with low computational overhead. The regressor comprised an ensemble of gradient-boosted decision trees with a maximum depth of 7, 31 leaves per tree, a learning rate of 0.1, and a minimum leaf size of 20 samples. Instance and feature subsampling ratios were both set to 0.8 to reduce overfitting. Separate models were trained for SBP and DBP.

\textbf{Training procedure:}  
Model training followed a two-stage strategy. First, the CNN feature extractor was trained on beat-level PPG samples using the Smooth L1 (Huber) loss \cite{girshick2015fastrcnn, gupta2020robust} and the Adam optimizer\cite{rstransformer,bock2018adam}, with an initial learning rate of $1\times10^{-3}$, weight decay of $1\times10^{-4}$, and a batch size of 256. Training was conducted for up to 60 epochs with early stopping based on validation mean absolute error (MAE). After convergence, CNN parameters were frozen, and the extracted embeddings were used to train the LightGBM regressors. Boosting was performed with early stopping and a maximum of 100 iterations.
\section{Results}
This section reports the BP estimation performance of the proposed hybrid CNN+LightGBM framework using synchronized ECG and PPG signals. Model performance was evaluated using both a large public dataset and a custom wearable dataset, and compared against baseline modeling approaches.


\subsection{Blood Pressure Estimation Performance}

To evaluate robustness and generalization, we conducted 30 independent experiments with different random subject-level splits. In each experiment, data from 80 subjects were randomly sampled from the PulseDB dataset, while data from 7 subjects were randomly selected from the wearable dataset. Subject-disjoint partitioning was strictly enforced across all splits to prevent information leakage between training and evaluation.

Table~\ref{tab:bp_performance} summarizes the prediction performance aggregated across all experiments. The proposed method achieved a mean absolute error (MAE) of $4.02 \pm 0.21$~mmHg for SBP and $1.79 \pm 0.05$~mmHg for DBP. The corresponding root mean squared errors (RMSE) were $10.79 \pm 0.93$~mmHg for SBP and $2.83 \pm 0.06$~mmHg for DBP. The low variance across repeated experiments indicates stable and reproducible performance.

\vspace{-0.5cm}
\begin{table}[htbp]
\centering
\caption{BP Prediction Performance over 30 Independent Runs}
\label{tab:bp_performance}
\begin{tabular}{lccc}
\toprule
\textbf{Metric} & \textbf{SBP} & \textbf{DBP} \\
\midrule
MAE (mean $\pm$ std)   & $4.02 \pm 0.21$ & $1.79 \pm 0.05$ \\
MAE range             & [3.60, 4.42]   & [1.68, 1.90] \\
RMSE (mean $\pm$ std) & $10.79 \pm 0.93$ & $2.83 \pm 0.06$  \\
\bottomrule
\end{tabular}
\end{table}

MAE was selected as the primary evaluation metric due to its clinical interpretability and consistency with international BP assessment standards, while RMSE was additionally reported to reflect sensitivity to larger prediction errors. The best-performing experimental run achieved an SBP MAE of 3.60~mmHg and a DBP MAE of 1.78~mmHg. This result represents the upper bound of performance observed across all evaluated splits.

\subsection{Stability and Generalization Analysis}
\begin{figure}
     \centering
     \includegraphics[width=1.00\linewidth]{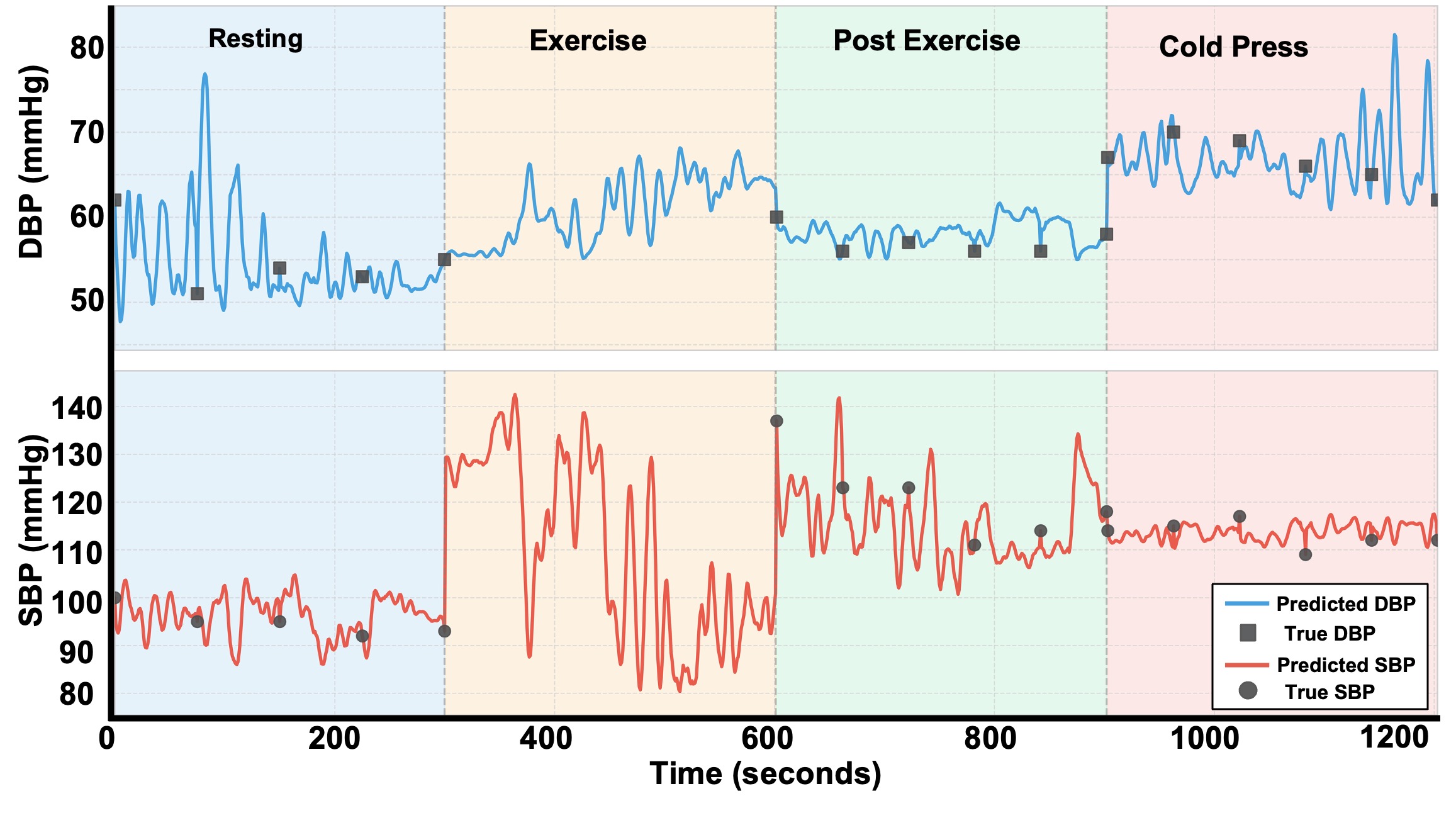}
     \caption{Predicted and reference SBP and DBP for a representative subject during resting, exercise, post-exercise recovery, and cold pressor phases.}
     \label{Fig: prediction_vs_actual}
 \end{figure}

Figure~\ref{Fig: prediction_vs_actual} shows representative continuous SBP and DBP predictions for a single subject across multiple physiological stages, including rest, exercise, post-exercise recovery, and cold pressor stress in Section \ref{subsec:protocol}. Discrete markers denote intermittent cuff-based reference measurements, while shaded regions indicate different experimental phases.

The proposed model captures stage-dependent BP dynamics and tracks temporal trends during rapid physiological transitions, including SBP/DBP elevation during exercise and cold exposure, followed by gradual recovery post exercise. Despite increased beat-to-beat variability during stress conditions, the predictions remain temporally aligned with the cuff-based references and preserve the expected directionality of BP changes, indicating subject-level generalization without stage-specific recalibration. Notably, during the exercise phase (highest motion and autonomic variability), the ear-PPG signal quality remained sufficient for beat-level inference, and the model consistently preserved the expected SBP increase, supporting the feasibility of the current ear-clip fixation approach (including adhesive reinforcement) for dynamic recordings. 

Across participants, similar trend-tracking behavior was observed, with SBP exhibiting larger and more consistent excursions across stages than DBP. This observation is consistent with established exercise physiology, in which SBP increases with workload while DBP typically shows smaller, more variable changes due to competing effects of cardiac output and peripheral vasodilation. The comparatively smaller dynamic range of DBP also renders it more sensitive to residual beat-level variability under motion, an expected characteristic of single-beat inference rather than an indication of systematic bias.

\begin{figure}
     \centering
     \includegraphics[width=1.00\linewidth]{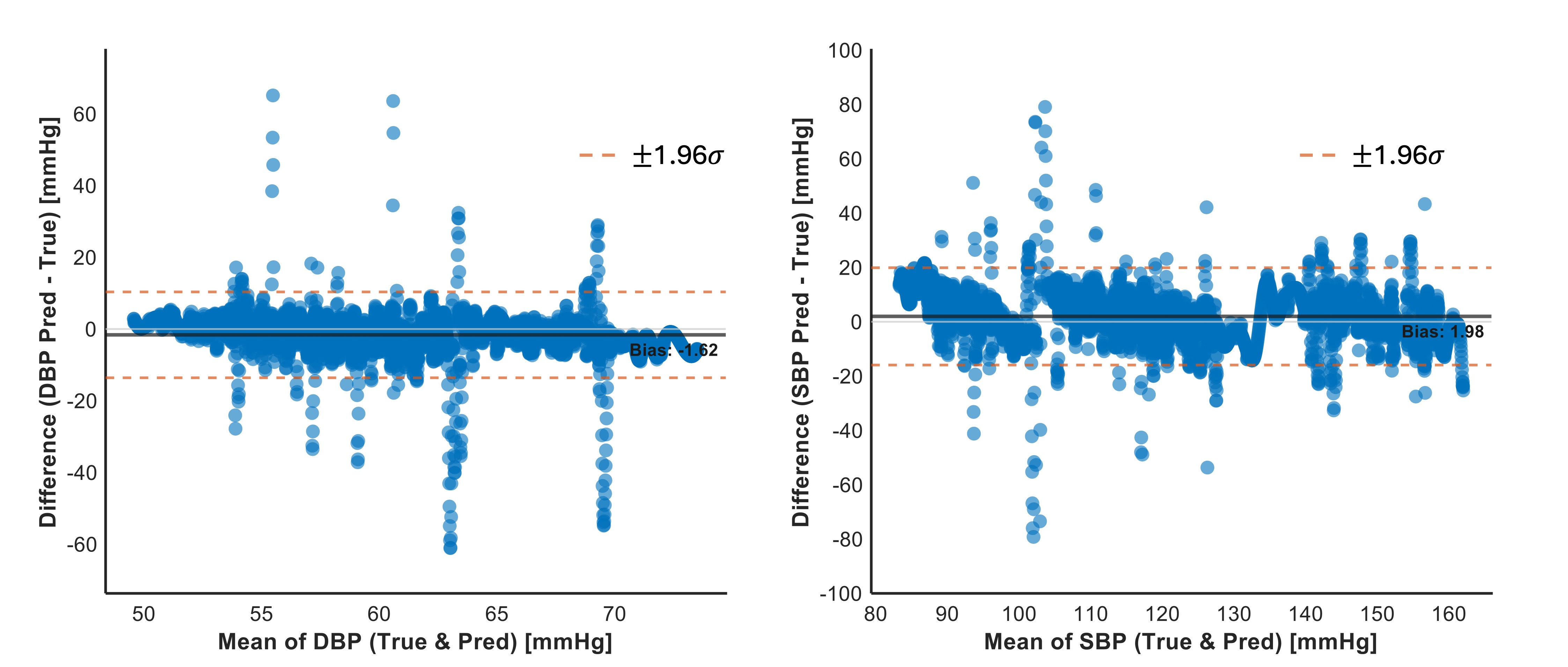}
     \caption{Bland--Altman plots for diastolic (left) and systolic (right) blood pressure estimates, showing prediction error (Pred-True) versus the mean of predicted and reference values.}
     \label{fig:performance_comparison}
 \end{figure}

Bland--Altman analysis was performed to assess agreement between predicted and reference BP values (Fig.~\ref{fig:performance_comparison}). For visualization clarity, a small fraction of extreme outliers was excluded from the plot only to reduce overplotting; however, all reported bias and limits of agreement were computed using the full test dataset. The analysis demonstrates strong agreement, with small mean bias (e.g., $-1.62$~mmHg for DBP) and relatively narrow 95\% limits of agreement, and no evident systematic error trends across the measured BP range.

The vertical clustering observed in the Bland--Altman plots arises from the discontinuous nature of cuff-based reference measurements. Multiple beat-level predictions generated within a short time window are compared against a single reference reading, resulting in repeated reference values along the horizontal axis. This phenomenon arises from the temporal mismatch between continuous predictions and intermittent ground truth rather than from model instability. Together, these results indicate that the proposed framework achieves stable, trend-consistent BP estimation across subjects and physiological states while maintaining clinically meaningful agreement with standard cuff measurements.


\subsection{Comparison of Modeling Approaches}
In addition to predictive accuracy, the proposed framework demonstrates high computational efficiency. The hybrid model comprises 152{,}802 trainable CNN parameters and 12{,}400 decision nodes in the LightGBM regressors, with an overall footprint below 2~MB. End-to-end inference requires approximately 3~ms per beat on GPU hardware, enabling throughput exceeding 300{,}000 beats/s. This efficiency is substantially higher than end-to-end deep learning approaches that rely on extended temporal windows and sequence processing.

To assess mobile deployment feasibility, the CNN feature extractor was converted from PyTorch to a Core ML–compatible format. The resulting model occupies only 0.58~MB and was evaluated on Apple silicon hardware to assess inference efficiency under realistic deployment conditions. Core ML benchmarking indicates a mean inference latency of 0.153~ms per beat (P95: 0.205~ms) with peak memory usage of 97.9~MB, and batch throughput exceeding 9{,}500 beats/s. These results indicate that the proposed beat-level feature extractor is compatible with execution on mobile gateways (e.g., iOS), with the tree-based regressor executed on the CPU.

To quantify the contribution of individual modeling components, an ablation study compared three approaches: (i) LightGBM using only hand-crafted physiological features, (ii) a CNN-only regression model operating on single-beat PPG waveforms, and (iii) the proposed hybrid CNN+LightGBM framework. Combined MAE (mmHg), defined as the sum of SBP and DBP MAE, was used as an aggregate error metric. As summarized in Table~\ref{tab:model_comparison}, the hybrid model consistently outperformed both single-branch baselines, achieving a 28.2\% reduction in combined MAE relative to the LightGBM-only model. In contrast, the CNN-only model exhibited substantially degraded performance, indicating that beat-level waveform morphology alone is insufficient for reliable BP estimation without explicit physiological context.

These results highlight the complementary roles of physiology-grounded features (e.g., PTT and HRV) and CNN-derived morphological embeddings. While hand-crafted features encode established cardiovascular mechanisms, the CNN captures subtle beat-level waveform characteristics. Together, this hybrid, single-beat design enables accurate and robust cuffless BP estimation with low computational overhead, supporting the practical feasibility of on-device BP monitoring under typical wearable constraints on battery, latency, and memory.
\begin{table}[t]
\centering
\caption{Ablation study comparing LightGBM with hand-crafted features, CNN-only regression using PPG beats, and the proposed hybrid LightGBM+CNN model.}
\label{tab:model_comparison}
\setlength{\tabcolsep}{3pt}
\renewcommand{\arraystretch}{1.05}
\footnotesize
\begin{tabular}{lcccc}
\toprule
\textbf{Model} & \textbf{SBP} & \textbf{DBP} & \textbf{Combined} & \textbf{Improv. (\%)} \\
\midrule
LightGBM        & 5.49  & 2.00 & 7.49  & -- \\
CNN             & 71.48 & 6.28 & 77.76 & $-938.0$ \\
LightGBM + CNN  & \textbf{3.60} & \textbf{1.78} & \textbf{5.38} & \textbf{+28.2} \\
\bottomrule
\end{tabular}
\vspace{-0.5cm}
\end{table}
These results demonstrate the complementary roles of physiology-grounded features (e.g., PTT and HRV) and CNN-derived morphological embeddings. While hand-crafted features encode established cardiovascular mechanisms, the CNN captures subtle beat-level waveform characteristics. Together, this hybrid, single-beat design enables accurate and robust cuffless BP estimation with low computational overhead. By avoiding long temporal windows, the pipeline reduces buffering and sequence-processing costs while preserving accuracy, supporting the practical feasibility of real-time, on-device BP monitoring under typical wearable constraints on battery, latency, and memory.


\section{Conclusion and Discussion}
In this study, we presented a multi-modal wearable framework for continuous, cuffless BP monitoring based on a hybrid CNN+LightGBM architecture. Using synchronized ECG and ear-PPG signals, the proposed approach achieved mean absolute errors of $4.02 \pm 0.21$~mmHg for systolic BP and $1.79 \pm 0.05$~mmHg for diastolic BP across 30 independent subject-level runs, satisfying established clinical accuracy criteria. These results demonstrate robust BP estimation performance under dynamic physiological conditions, including elevated activity and autonomic stress, with accurate tracking of stage-dependent BP trends despite beat-level variability.

A key strength of the proposed framework lies in its physiology-guided feature fusion strategy. By combining domain-specific features (e.g., PTT and HRV) with CNN-derived single-beat PPG morphological embeddings, the model achieves accurate BP estimation without reliance on long temporal windows or computationally intensive sequence models. This design is particularly advantageous for wearable deployment, where latency, power consumption, and memory resources are tightly constrained.

Across subjects and experimental stages, systolic BP exhibited larger and more consistent excursions—particularly during exercise while diastolic BP changes were smaller and more heterogeneous, consistent with established cardiovascular physiology and prior exercise studies. The smaller dynamic range of DBP renders it inherently more sensitive to residual beat-level variability, an expected characteristic of single-beat inference rather than an indication of systematic estimation error.

Several limitations of the current implementation warrant discussion. First, although the ear-PPG hardware supports four simultaneous optical wavelengths, this study utilized a single optimal channel to minimize model complexity. Multi-wavelength fusion may further improve robustness to motion artifacts and wavelength-dependent perfusion changes and will be explored in future work. Second, while the combined dataset spans 109 subjects, additional validation on larger and more diverse populations is necessary to fully assess generalizability. Third, all signals were resampled to 125~Hz, which may limit the temporal precision of features such as PTT compared to higher sampling rates. Finally, the current beat-wise processing framework does not explicitly model long-term temporal dependencies; future work may investigate lightweight temporal extensions to improve performance during abrupt physiological transitions. The current ear-PPG hardware was adapted from a head-mounted design and is larger than necessary for earlobe-specific integration. Future miniaturized designs with optimized optical layouts are expected to improve mechanical coupling and motion robustness, and the trade-offs between reflective and transmissive ear-based PPG will be explored.

In this study, beat-level PPG segments were extracted using a fixed temporal window selected based on empirical tuning to balance morphological fidelity and computational efficiency. While this configuration performed consistently across datasets, future deployment-oriented work may further refine windowing strategies based on target hardware constraints, latency requirements, and application-specific operating conditions.

Overall, this work demonstrates that integrating data-driven morphological representations with physiology-grounded features provides a reliable, efficient, and practically deployable solution for continuous cuffless BP monitoring. The proposed hybrid framework offers a strong foundation for next-generation wearable cardiovascular monitoring systems in ambulatory and real-world settings.

\section*{Acknowledgments}
The authors thank Prof. John A. Rogers for providing access to the wearable fNIRS sensors used in this study.




\bibliographystyle{IEEEtran}
\bibliography{references}

\vspace{12pt}

\end{document}